\pdfoutput=1

\documentclass[11pt]{article}

\usepackage[]{acl}
\usepackage{amsmath}
\usepackage{times}
\usepackage{latexsym}
\usepackage{array}
\usepackage{tabularx}
\usepackage{multirow}
\usepackage{multicol}
\usepackage{comment}
\usepackage{booktabs}
\usepackage{graphicx}
\usepackage{url}
\usepackage{ulem}
\usepackage{xcolor}
\usepackage{wasysym}
\usepackage{marvosym}
\usepackage{subcaption}
\usepackage{caption}

\newcolumntype{P}[1]{>{\centering\arraybackslash}p{#1}}

\definecolor{scarlet}{rgb}{0.8, 0.0, 0.0}
\definecolor{purple}{rgb}{0.41, 0.21, 0.61}

\usepackage[T1]{fontenc}

\usepackage[utf8]{inputenc}

\usepackage{microtype}

\usepackage{inconsolata}

\usepackage{xspace}

\usepackage{xcolor}
\newcommand{\multilingualholisticbias}{\textsc{MultilingualHolisticBias}\xspace}

\newcommand{\customfigref}[2]{%
  \hyperlink{#1}{~\ref*{#1}#2}%
}
\usepackage{mxedruli}
\usepackage{dingbat, xcolor}

\usepackage{placeins}
\usepackage{adjustbox}

\title{Gender-specific Machine Translation with Large Language Models}

\author{Eduardo Sánchez$^{\text{$\alpha$ $\beta$}}$ \quad Pierre Andrews$^{\text{$\alpha$}}$ \quad Pontus Stenetorp$^{\text{$\beta$}}$ \\ 
\bf Mikel Artetxe$^{\text{$\gamma$}}$ \quad Marta R. Costa-jussà$^{\text{$\alpha$}}$ \\
$^{\text{$\alpha$}}$Meta \quad
$^{\text{$\beta$}}$UCL Centre for Artificial Intelligence  \\
$^{\text{$\gamma$}}$University of the Basque Country (UPV/EHU) \\
\texttt{\{eduardosanchez, mortimer, costajussa\}@meta.com} \\
\texttt{p.stenetorp@cs.ucl.ac.uk} \quad \texttt{mikel.artetxe@ehu.eus} \\
}

\begin{document}
\maketitle
\begin{abstract}
While machine translation (MT) systems have seen significant improvements, it is still common for translations to reflect societal biases, such as gender bias. Decoder-only Large Language Models (LLMs) have demonstrated potential in MT, albeit with performance slightly lagging behind traditional encoder-decoder Neural Machine Translation (NMT) systems. However, LLMs offer a unique advantage: the ability to control the properties of the output through prompts. In this study, we leverage this flexibility to explore LLaMa's capability to produce gender-specific translations. Our results indicate that LLaMa can generate gender-specific translations with translation accuracy and gender bias comparable to NLLB, a state-of-the-art multilingual NMT system. Furthermore, our experiments reveal that LLaMa's gender-specific translations rely on coreference resolution to determine gender, showing higher gender variance in gender-ambiguous datasets but maintaining consistency in less ambiguous contexts. This research investigates the potential and challenges of using LLMs for gender-specific translations as an instance of the controllability of outputs offered by LLMs.
\end{abstract}

\section{Introduction}

\begin{figure}[ht]
  \centering
  \begin{subfigure}[b]{0.46\textwidth}
    \centering
    \fbox{\includegraphics[width=\textwidth, trim={2cm 20cm 6cm 2cm}, clip]{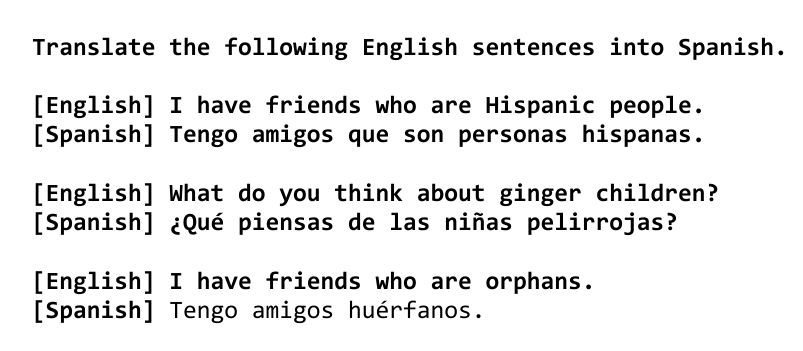}}
    \caption{Standard MT template.}
    \label{fig:simple_prompt}
  \end{subfigure}
  \hfill
  \begin{subfigure}[b]{0.46\textwidth}
    \centering
    \fbox{\includegraphics[width=\textwidth, trim={2cm 19cm 6cm 2cm}, clip]{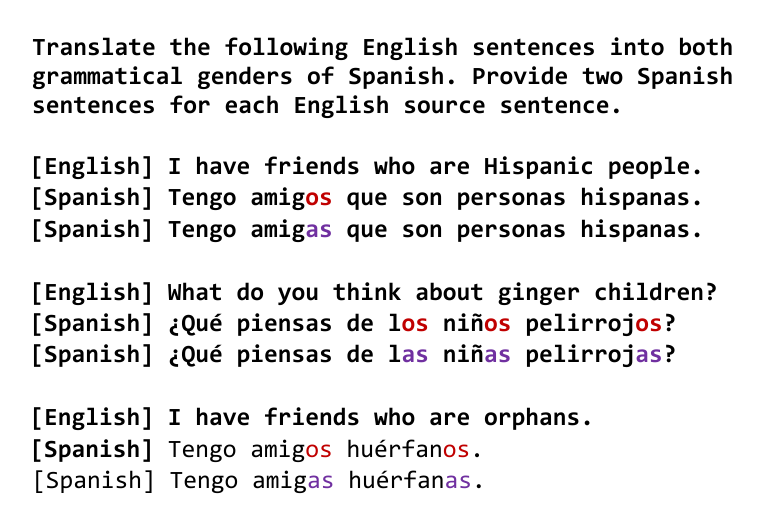}}
    \caption{Gender-specific template.}
    \label{fig:double_prompt}
  \end{subfigure}
  \caption{Prompt templates employed. Languages like Spanish (displayed here) present morphological differences between masculine (\textcolor{scarlet}{red}) and female (\textcolor{purple}{purple}) grammatical genders.}
  \label{fig:both_gens}
\end{figure}

Over the last few years, machine translation (MT) systems have seen significant improvements with the introduction of Neural Machine Translation (NMT). Despite these advances, MT can reflect societal biases, such as gender bias.
A prominent instance of this problem occurs when the target language marks the grammatical gender, but the source language does not (Fig. \ref{fig:double_prompt}). In such instances, translating into either gender can be correct, but MT systems tend to pick the gender that corresponds to stereotypical associations (e.g., associating certain professions to males and others to females \cite{escude-font-costa-jussa-2019-equalizing}). Instead, it would be preferable to generate both options, and/or let the user control the gender.

Decoder-only Large Language Models (LLMs) have shown MT capabilities inferior to but competitive with encoder-decoder Neural Machine Translation (NMT) systems \cite{chowdhery2022palm, agrawal-etal-2023-context, zhang2023prompting, bawden2023investigating, zhu2023multilingual, jiao2023chatgpt, hendy2023good}. However, LLMs have been proven to allow for more control over the properties of the output \cite{garcia2023unreasonable, moslem2023adaptive, pilault2023interactivechainprompting}. While NMT models are trained to accept a single sequence and output its translation, LLMs make it possible to condition the output format with a prompt.

The task of inferring gender-specific translations from a gender-neutral source has been addressed mainly through post-editing, the most popular solution being Google Translate's post-translation gender rewriter \cite{Johnson_2020}. The proposed system produces a single sentence that is then switched into the opposite gender using a second language-specific model. This approach is limited by having to train language-specific gender-switching models and the breadth of patterns it can cover.

Given the flexibility of prompting, we explore the capacity of LLMs to produce gender-specific translations for languages with grammatical gender from gender-neutral sources without significant losses in translation quality or increases in gender bias. 

We use in-context examples (ICEs) to elicit the task of translation from a gender-neutral source to two gender-specific targets (Figure \ref{fig:double_prompt}).
Additionally, we evaluate the quality of the gender-specific translations on two aspects: gender bias (measured against coreference resolution accuracy) and translation quality (measured in BLEU).

We show that it is possible to generate gender-specific translations with translation quality and gender bias competitive with NLLB, with a slightly better performance than LLaMa for masculine/both references evaluation and over 10 BLEU points for the feminine reference. We also demonstrate the reliance on coreference resolution of the gender-specific translation method, showing steep decreases in performance when using the opposite gender as an evaluation reference in a gender-focused dataset (\multilingualholisticbias{}), but exhibiting lesser variance in a general translation dataset (FLoRes).

\section{Related Work} \label{sec:related-work}

\paragraph{MT and controlled output with LLMs}

A few papers have evaluated the quality of MT using different models and GPT-based commercial products, such as PALM \cite{chowdhery2022palm}, XGLM \cite{agrawal-etal-2023-context}, GLM \cite{zhang2023prompting}, BLOOM \cite{bawden2023investigating}, OPT \cite{zhu2023multilingual} or ChatGPT \cite{jiao2023chatgpt,hendy2023good}. They conclude that the translation quality comes close but remains behind the performance of NMTs \cite{kocmi-etal-2023-findings}. Using LLMs can, however, allow for more control over the properties of the output without further finetuning, such as specifying the language variety and style of the translation \cite{garcia2023unreasonable}, producing terminology-constrained translations \cite{moslem2023adaptive} or using an iterative prompting process to clarify ambiguities in the source sentence \cite{pilault2023interactivechainprompting}. Challenges persist in the area of hallucinations \cite{zhang2023prompting,guerreiro2023hallucinations} and in performance in low-resource languages \cite{bawden2023investigating,zhu2023multilingual}.

\paragraph{Gender Bias in MT}

Some authors have worked in analyzing and mitigating gender bias in MT. \citet{DBLP:journals/corr/abs-1809-02208} studied the bias of the commercial translation system Google Translate and found that it yields male defaults much more frequently than what would be expected from US demographic data. \citet{Costa-jussà_Escolano_Basta_Ferrando_Batlle_Kharitonova_2022} investigate the role of model architecture in the level of gender bias, while \citet{mechura-2022-taxonomy} looks at the source sentences and elaborates a taxonomy of the features that induce gender bias into the translations. Others have looked more closely at the challenge of gender bias mitigation. \citet{stafanovics-etal-2020-mitigating} assume that it's not always possible to infer all the necessary information from the source sentence alone and a method that uses word-level annotations containing information about the subject’s gender to decouple the task of performing an unbiased translation from the task of acquiring gender-specific information. \citet{saunders-byrne-2020-reducing} treat the mitigation as a domain adaptation problem, using transfer learning on a small set of trusted, gender-balanced examples to achieve considerable gains with a fraction of the from-scratch training costs. \citet{fleisig2022mitigating} develop a framework to make NMT systems suitable for gender bias mitigation through adversarial learning, adjusting the training objective at fine-tuning time. Finally, \citet{wang-etal-2022-measuring} focus on existing biases in person name translation, applying a data augmentation technique consisting of randomly switching entities, obtaining satisfactory results.

\section{Experimental Framework}

\paragraph{Data}
For our main experiments, we use the \multilingualholisticbias{} dataset \cite{costajussà2023multilingual}, a multilingual subset of Holistic Bias \cite{smith-etal-2022-im} with separate translations for each noun class or grammatical gender for those languages that make use of them\footnote{For this study, we selected the subset of languages that make use of grammatical genders or noun classes and for which there is correlation between grammatical gender and natural gender, allowing us to establish a relationship between gender bias and the accuracy of coreference resolution in a model.}. An example of an entry of the  dataset can be found in Table \ref{tab:mhb_examples} (Appendix \ref{app:mhb}). We also filtered out the languages which are not explicitly present in the LLaMa-2 pre-training set \cite{touvron2023llama2}. Since MHB was created translating a limited number of templates, we exclude entries with a similar template when performing ICL. A complete list of languages used from the \multilingualholisticbias{} dataset can be found in Appendix \ref{app:langs}. Additionally, we use a subset of BUG's \cite{levy-etal-2021-collecting-large} gold (human-annotated) set for gender bias analysis and the FLoRes \cite{nllb2022, goyal2021flores, guzman2019two} devtest set to reproduce our results in the general domain.

\paragraph{Models} We use LLaMA-2 \cite{touvron2023llama2}, a decoder-only model, and NLLB \cite{nllb2022}, an encoder-decoder model. We use the NLLB-200 version with 3 billion parameters. For LLaMa-2 we use the 70 billion parameter version. We prompt LLaMa-2 with ICEs (Figure \ref{fig:double_prompt}) to elicit the gender-specific translation task.  To facilitate comparisons, we also prompt LLaMa-2 with a standard MT in-context learning (ICL) prompt template (Figure \ref{fig:simple_prompt}).

\paragraph{Evaluation}
Following the work of \citet{costajussà2023multilingual}, we use the sacrebleu implementation of spBLEU \cite{flores101} to compute the translation quality with \texttt{`add-k = 1'} smoothing. We also provide evaluations in chrF \cite{popovic-2015-chrf}, COMET \cite{rei2020comet}, BLEURT \cite{sellam-etal-2020-bleurt} and BLASER \cite{chen-etal-2023-blaser} as alternative metrics. For gender bias evaluation, we use \citet{stanovsky-etal-2019-evaluating}'s reference-less coreference resolution metric.

\paragraph{Experimental Setup}
We investigate the capability of LLaMa to produce gender-specific translations. We prompt LLaMa with 8 ICEs conformed by source, masculine and feminine translations from \multilingualholisticbias{} (Fig. \ref{fig:double_prompt}). We also prompt LLaMa with a standard MT template, randomly selecting among the available translations when there's more than one option (Fig. \ref{fig:simple_prompt}). Hereinafter all experiments are performed with these settings. For NLLB, we calculate three BLEU scores on the output: one with the masculine reference, one with the feminine reference and one with both. In the case of LLaMa, we calculate two BLEU scores for each gender-specific output: one with the corresponding gender's reference and one with both references, for a total of four BLEU scores per generation.

\section{Results}

\begin{table}[t!]
\centering
\small
\setlength{\tabcolsep}{4pt} 

\begin{tabular}{llclll}
\toprule
&& masc & fem & both  \\
\midrule
NLLB & unsp & 40.07 & 28.67 & 40.41 \\
\midrule
\multirow{3}{*}{\textbf{LLaMA}}
& unsp & 41.57 & 30.92 & 42.43 \\
& masc & \textbf{41.63} & \textcolor{gray}{30.12} & 42.08 \\
& fem & \textcolor{gray}{31.84} & \textbf{39.55} & \textbf{43.37} \\
\bottomrule
\end{tabular}

\caption{BLEU scores of the unspecified, masculine and feminine outputs of NLLB and LLaMa evaluated on masculine, feminine, and both references of \multilingualholisticbias{}}
\label{tab:main_results}
\end{table} 

\paragraph{Gender-specific MT results in \multilingualholisticbias{}} As Table \ref{tab:main_results} shows, on average LLaMa outperforms NLLB on all three references. While the differences between masculine/both references are moderate (Figs.\customfigref{fig:bleu_scores}{a} \&\customfigref{fig:bleu_scores}{c} in Appendix \ref{app:full_res}), LLaMa outperforms NLLB by an average of over 10 BLEU points for the feminine reference (Fig.\customfigref{fig:bleu_scores}{b} in Appendix \ref{app:full_res}), highlighting the capacity of gender-specific MT to provide comparable results for masculine and feminine outputs. We also assessed the capacity of the BLEU evaluation metric to capture gender nuances. We inverted masculine and feminine references and found steep decreases, indicating the effectiveness of our approach in specifying the grammatical gender. Additionally, we provide evaluations in chrF, COMET, BLEURT and BLASER, which show consistency with BLEU scores. Full results can be found in Appendix \ref{app:full_res}.


\paragraph{Gender bias MT results in BUG}
Besides translation accuracy, we're interested in verifying the incidence of gender bias in gender-specific translations with respect to unspecified translation. We translate BUG's gold set, reusing \multilingualholisticbias{} examples for ICL. BUG's gold set is made of English sentences that require unambiguous coreference resolution or grammatical gender utilization to produce correct translations, regardless of stereotypical associations. To ensure fairness in our analysis, we sampled four same-sized subsets from BUG gold, each corresponding to a combination of stereotypical/antistereotypical correferences and male/female nouns. \citet{stanovsky-etal-2019-evaluating} and \citet{levy-etal-2021-collecting-large} found that several (encoder-decoder) NMTs are significantly prone to translate based on gender stereotypes rather than more meaningful context. We verify to which degree these errors are reproduced by LLaMa in gender-specific translations. 
When performing the translation of BUG, we noticed that the phenomenon of empty or incomplete outputs occasionally occurs (i.e., either only one output or no output at all is produced). Since a gender bias analysis is not defined over an empty sentence, for each language we evaluate all models in the subset that has been correctly generated by LLaMa both in the unspecified and the gender-specific modalities.

\begin{table}[t!]
\centering
\small
\renewcommand{\arraystretch}{1.2} 
\setlength{\tabcolsep}{1pt} 

\begin{tabular}{ccc|cc|cc|cc}
\toprule
                 & \multicolumn{2}{c}{NLLB} & \multicolumn{6}{c}{LLaMa} \\
\cmidrule(lr){2-3} \cmidrule(lr){4-9}
                & \multicolumn{2}{c}{unsp} &\multicolumn{2}{c}{unsp}  & \multicolumn{2}{c}{masc}&  \multicolumn{2}{c}{fem} \\
\midrule
      & acc.($\uparrow$) & $\Delta_B$($\downarrow$) & acc.($\uparrow$) & $\Delta_B$($\downarrow$) & acc. & $\Delta_B$($\downarrow$) & acc.($\uparrow$) & $\Delta_B$ ($\downarrow$)   \\
\midrule      
ces   & 59.3 & \underline{6.5}  & 57.2 & 11.3 & \textbf{61.7} & 10.1 & 48.4 & 8.8 \\
deu   & 66.4 & 11.8 & 67.8 & 10.8 & \textbf{70.6} & 9.5 & 52.4 & \underline{8.6} \\
ita   & 46.2 & \underline{12.5} & 45.4 & 13.7 & \textbf{46.5} & 14.4 & 38.9 & 14.2 \\
spa   & \textbf{52.5} & \underline{10.1} & 50.0 & 11.4 & 49.4 & 14.4 & 34.2 & 29.4 \\
rus   & 36.6 & 25.0 & \textbf{39.5} & 23.8 & 38.1 & 27.5 & 36.9 & \underline{16.7} \\
ukr   & 41.2 & 11.1 & 42.1 & 10.1 & \textbf{43.2} & 8.8 & 39.0 & \underline{1.0} \\
\bottomrule
\end{tabular}
\caption{Noun gender prediction accuracy on the subset of BUG's gold dataset's fully generated gender-specific translations with LLaMa, compared to NLLB's prediction accuracy. LLaMa results are presented for male (m.), female (f.), and unspecified (unsp.) genders. We also show the differences in accuracy between male nouns and female nouns for each case ($\Delta_B$)}
\label{tab:bug}
\end{table}

Table \ref{tab:bug} shows that LLaMa's masculine output's noun gender prediction accuracy outperforms NLLB's for almost every language, but underperforms NLLB for feminine outputs. Difference of accuracy between genders for the same type of output ($\Delta_B$) is comparable accross models. 

\paragraph{General domain MT results in FLoRes} A possible concern about previous results is that they are produced by the system forcing a specific gender instead of performing coreference resolution to determine the correct gender. To study whether this is the case, we assess the difference in performance for each produced gender when there aren't major gender ambiguities to translate. In this case, a robust model should not have significant differences between both genders. We translate FLoRes's devtest set into ten languages included in LLaMa's training corpus. Given that FLoRes is a general domain dataset, ambiguities should not be prevalent and both outputs should tend to converge. We use \multilingualholisticbias{} as ICEs and compare the BLEU scores of both outputs. The list of languages we translate into for this experiment can be found in Table \ref{tab:langs} (Appendix \ref{app:langs}).

The results show minor differences between both genders, suggesting a coreference resolution-based gender-specific generation rather than on mechanically switching the grammatical gender of the words of the sentence.

\begin{table}
\centering
\tiny 
\renewcommand{\arraystretch}{1.2} 
\setlength{\tabcolsep}{2pt} 

\begin{tabular}{l|cccccccccc|l}
\toprule
& \text{cat} & \text{deu} & \text{fra} & \text{ita} & \text{nld} & \text{por} & \text{rus} & \text{spa} & \text{swe} & \text{ukr} & \text{avg} \\
\midrule
\text{nllb} & 45.81 & 43.38 & 53.43 & 36.34 & 33.96 & 53.05 & 38.40 & 32.99 & 47.58 & 36.31 & 42.13 \\
\text{unsp.} & 46.05 & 41.79 & 52.24 & 34.70 & 32.54 & 51.76 & 36.17 & 31.34 & 47.74 & 36.02 & 41.04 \\
\text{masc.} & 46.06 & 42.18 & 52.05 & 34.46 & 32.36 & 51.68 & 36.23 & 31.25 & 47.90 & 36.05 & 41.02 \\
\text{fem.} & 43.83 & 41.02 & 50.25 & 33.25 & 31.43 & 49.29 & 34.57 & 29.72 & 47.63 & 35.38 & 39.64 \\
\text{$\Delta_F$} & 2.23 & 1.16 & 1.80 & 1.21 & 0.93 & 2.39 & 1.66 & 1.53 & 0.27 & 0.67 & 1.39 \\
\bottomrule
\end{tabular}
\caption{BLEU scores for each output of LLaMa's gender-specific translation on FLoRes's testset. $\Delta_F$ denotes the difference between male and female translations. Since FLoRes's sentences are not expected to contain a high rate of ambiguity, a correct translation should tend to be identical in both outputs.}
\label{tab:flores-gender}
\end{table}

\section{Conclusions} \label{sec:conclusions}

In this paper, we explored the capabilities and limitations a decoder-only LLM to produce gender-specific translations. We observed that LLaMa's gender-specific translations' accuracy is consistently above NLLB's.  We also showed that LLaMa's gender-specific translations' gender bias is comparable to NLLB's. These results indicate that it is possible to use LLMs to produce gender-specific translations without compromising on lower translation accuracy or higher gender bias. Our experiments also reveal that LLaMa's translations rely on coreference resolution to determine gender, showing significant performance drops when evaluated against opposite-gender references in gender-ambiguous datasets, but maintaining consistency in less ambiguous contexts.

While these results are promising indicator of the flexibility of the output in the task of MT for languages present in LLaMa's training set, the limited multilinguality of currently available LLMs limits the application of this approach to a subset of the languages present in state-of-the-art NMT models. More work is needed to bring LLMs' multilingual capabilities on par with NMTs.

\section*{Limitations}

Even though we performed a diverse set of experiments, some limitations arise due to the vastness of the research space we're dealing with. The study heavily relies on the effectiveness of prompt engineering, specifically in providing accurate ICEs. The conclusions drawn are thus constrained by the quality and relevance of the prompts used. Variations in prompt structure or content could yield different results. Moreover, the study focuses on a particular model, LLaMa-2, leaving out an exploration of alternative LLMs that could yield different results.

\multilingualholisticbias{}'s small number of templates and their simplicity limit the scope of our results. An exploration with a more diverse dataset could bring additional insights to our conclusions.

\section*{Ethics Statement}

The understanding of nuanced gender contexts is intricate and can be challenging even for humans. The study tends to approach gender in a binary manner, which might not account for social perceptions among some of the users of these languages. This limitation is inherent in the current state of the field and warrants future investigations into better representation and handling of gender-related nuances.

Furthermore, the stereotypical and non-stereotypical datasets were built based on the US Department of Labor data. Since we work with a variety of world languages, the proportions stated on these datasets might not reflect the realities of the users of the wide range of languages employed in this study.

\bibliography{anthology,custom}

\begin{thebibliography}{34}
\expandafter\ifx\csname natexlab\endcsname\relax\def\natexlab#1{#1}\fi

\bibitem[{Agrawal et~al.(2023)Agrawal, Zhou, Lewis, Zettlemoyer, and
  Ghazvininejad}]{agrawal-etal-2023-context}
Sweta Agrawal, Chunting Zhou, Mike Lewis, Luke Zettlemoyer, and Marjan
  Ghazvininejad. 2023.
\newblock \href {https://aclanthology.org/2023.findings-acl.564} {In-context
  examples selection for machine translation}.
\newblock In \emph{Findings of the Association for Computational Linguistics:
  ACL 2023}, pages 8857--8873, Toronto, Canada. Association for Computational
  Linguistics.

\bibitem[{Bawden and Yvon(2023)}]{bawden2023investigating}
Rachel Bawden and François Yvon. 2023.
\newblock \href {http://arxiv.org/abs/2303.01911} {Investigating the
  translation performance of a large multilingual language model: the case of
  bloom}.

\bibitem[{Chen et~al.(2023)Chen, Duquenne, Andrews, Kao, Mourachko, Schwenk,
  and Costa-juss{\`a}}]{chen-etal-2023-blaser}
Mingda Chen, Paul-Ambroise Duquenne, Pierre Andrews, Justine Kao, Alexandre
  Mourachko, Holger Schwenk, and Marta~R. Costa-juss{\`a}. 2023.
\newblock \href {https://aclanthology.org/2023.acl-long.504} {{BLASER}: A
  text-free speech-to-speech translation evaluation metric}.
\newblock In \emph{Proceedings of the 61st Annual Meeting of the Association
  for Computational Linguistics (Volume 1: Long Papers)}, pages 9064--9079,
  Toronto, Canada. Association for Computational Linguistics.

\bibitem[{Chowdhery et~al.(2022)Chowdhery, Narang, Devlin, Bosma, Mishra,
  Roberts, Barham, Chung, Sutton, Gehrmann, Schuh, Shi, Tsvyashchenko, Maynez,
  Rao, Barnes, Tay, Shazeer, Prabhakaran, Reif, Du, Hutchinson, Pope, Bradbury,
  Austin, Isard, Gur-Ari, Yin, Duke, Levskaya, Ghemawat, Dev, Michalewski,
  Garcia, Misra, Robinson, Fedus, Zhou, Ippolito, Luan, Lim, Zoph, Spiridonov,
  Sepassi, Dohan, Agrawal, Omernick, Dai, Pillai, Pellat, Lewkowycz, Moreira,
  Child, Polozov, Lee, Zhou, Wang, Saeta, Diaz, Firat, Catasta, Wei,
  Meier-Hellstern, Eck, Dean, Petrov, and Fiedel}]{chowdhery2022palm}
Aakanksha Chowdhery, Sharan Narang, Jacob Devlin, Maarten Bosma, Gaurav Mishra,
  Adam Roberts, Paul Barham, Hyung~Won Chung, Charles Sutton, Sebastian
  Gehrmann, Parker Schuh, Kensen Shi, Sasha Tsvyashchenko, Joshua Maynez,
  Abhishek Rao, Parker Barnes, Yi~Tay, Noam Shazeer, Vinodkumar Prabhakaran,
  Emily Reif, Nan Du, Ben Hutchinson, Reiner Pope, James Bradbury, Jacob
  Austin, Michael Isard, Guy Gur-Ari, Pengcheng Yin, Toju Duke, Anselm
  Levskaya, Sanjay Ghemawat, Sunipa Dev, Henryk Michalewski, Xavier Garcia,
  Vedant Misra, Kevin Robinson, Liam Fedus, Denny Zhou, Daphne Ippolito, David
  Luan, Hyeontaek Lim, Barret Zoph, Alexander Spiridonov, Ryan Sepassi, David
  Dohan, Shivani Agrawal, Mark Omernick, Andrew~M. Dai,
  Thanumalayan~Sankaranarayana Pillai, Marie Pellat, Aitor Lewkowycz, Erica
  Moreira, Rewon Child, Oleksandr Polozov, Katherine Lee, Zongwei Zhou, Xuezhi
  Wang, Brennan Saeta, Mark Diaz, Orhan Firat, Michele Catasta, Jason Wei,
  Kathy Meier-Hellstern, Douglas Eck, Jeff Dean, Slav Petrov, and Noah Fiedel.
  2022.
\newblock \href {http://arxiv.org/abs/2204.02311} {Palm: Scaling language
  modeling with pathways}.

\bibitem[{Costa-jussà et~al.(2023)Costa-jussà, Andrews, Smith, Hansanti,
  Ropers, Kalbassi, Gao, Licht, and Wood}]{costajussà2023multilingual}
Marta~R. Costa-jussà, Pierre Andrews, Eric Smith, Prangthip Hansanti,
  Christophe Ropers, Elahe Kalbassi, Cynthia Gao, Daniel Licht, and Carleigh
  Wood. 2023.
\newblock \href {http://arxiv.org/abs/2305.13198} {Multilingual holistic bias:
  Extending descriptors and patterns to unveil demographic biases in languages
  at scale}.

\bibitem[{Costa-jussà et~al.(2022)Costa-jussà, Escolano, Basta, Ferrando,
  Batlle, and
  Kharitonova}]{Costa-jussà_Escolano_Basta_Ferrando_Batlle_Kharitonova_2022}
Marta~R. Costa-jussà, Carlos Escolano, Christine Basta, Javier Ferrando, Roser
  Batlle, and Ksenia Kharitonova. 2022.
\newblock \href {https://doi.org/10.1609/aaai.v36i11.21442} {Interpreting
  gender bias in neural machine translation: Multilingual architecture
  matters}.
\newblock \emph{Proceedings of the AAAI Conference on Artificial Intelligence},
  36(11):11855--11863.

\bibitem[{Escud{\'e}~Font and
  Costa-juss{\`a}(2019)}]{escude-font-costa-jussa-2019-equalizing}
Joel Escud{\'e}~Font and Marta~R. Costa-juss{\`a}. 2019.
\newblock \href {https://doi.org/10.18653/v1/W19-3821} {Equalizing gender bias
  in neural machine translation with word embeddings techniques}.
\newblock In \emph{Proceedings of the First Workshop on Gender Bias in Natural
  Language Processing}, pages 147--154, Florence, Italy. Association for
  Computational Linguistics.

\bibitem[{Fleisig and Fellbaum(2022)}]{fleisig2022mitigating}
Eve Fleisig and Christiane Fellbaum. 2022.
\newblock \href {http://arxiv.org/abs/2203.10675} {Mitigating gender bias in
  machine translation through adversarial learning}.

\bibitem[{Garcia et~al.(2023)Garcia, Bansal, Cherry, Foster, Krikun, Feng,
  Johnson, and Firat}]{garcia2023unreasonable}
Xavier Garcia, Yamini Bansal, Colin Cherry, George Foster, Maxim Krikun,
  Fangxiaoyu Feng, Melvin Johnson, and Orhan Firat. 2023.
\newblock \href {http://arxiv.org/abs/2302.01398} {The unreasonable
  effectiveness of few-shot learning for machine translation}.

\bibitem[{Goyal et~al.(2021{\natexlab{a}})Goyal, Gao, Chaudhary, Chen, Wenzek,
  Ju, Krishnan, Ranzato, Guzm\'{a}n, and Fan}]{goyal2021flores}
Naman Goyal, Cynthia Gao, Vishrav Chaudhary, Peng-Jen Chen, Guillaume Wenzek,
  Da~Ju, Sanjana Krishnan, Marc'Aurelio Ranzato, Francisco Guzm\'{a}n, and
  Angela Fan. 2021{\natexlab{a}}.
\newblock The flores-101 evaluation benchmark for low-resource and multilingual
  machine translation.

\bibitem[{Goyal et~al.(2021{\natexlab{b}})Goyal, Gao, Chaudhary, Chen, Wenzek,
  Ju, Krishnan, Ranzato, Guzm\'{a}n, and Fan}]{flores101}
Naman Goyal, Cynthia Gao, Vishrav Chaudhary, Peng-Jen Chen, Guillaume Wenzek,
  Da~Ju, Sanjana Krishnan, Marc'Aurelio Ranzato, Francisco Guzm\'{a}n, and
  Angela Fan. 2021{\natexlab{b}}.
\newblock The flores-101 evaluation benchmark for low-resource and multilingual
  machine translation.

\bibitem[{Guerreiro et~al.(2023)Guerreiro, Alves, Waldendorf, Haddow, Birch,
  Colombo, and Martins}]{guerreiro2023hallucinations}
Nuno~M. Guerreiro, Duarte Alves, Jonas Waldendorf, Barry Haddow, Alexandra
  Birch, Pierre Colombo, and André F.~T. Martins. 2023.
\newblock \href {http://arxiv.org/abs/2303.16104} {Hallucinations in large
  multilingual translation models}.

\bibitem[{Guzm\'{a}n et~al.(2019)Guzm\'{a}n, Chen, Ott, Pino, Lample, Koehn,
  Chaudhary, and Ranzato}]{guzman2019two}
Francisco Guzm\'{a}n, Peng-Jen Chen, Myle Ott, Juan Pino, Guillaume Lample,
  Philipp Koehn, Vishrav Chaudhary, and Marc'Aurelio Ranzato. 2019.
\newblock Two new evaluation datasets for low-resource machine translation:
  Nepali-english and sinhala-english.

\bibitem[{Hendy et~al.(2023)Hendy, Abdelrehim, Sharaf, Raunak, Gabr,
  Matsushita, Kim, Afify, and Awadalla}]{hendy2023good}
Amr Hendy, Mohamed Abdelrehim, Amr Sharaf, Vikas Raunak, Mohamed Gabr, Hitokazu
  Matsushita, Young~Jin Kim, Mohamed Afify, and Hany~Hassan Awadalla. 2023.
\newblock \href {http://arxiv.org/abs/2302.09210} {How good are gpt models at
  machine translation? a comprehensive evaluation}.

\bibitem[{Jiao et~al.(2023)Jiao, Wang, tse Huang, Wang, and
  Tu}]{jiao2023chatgpt}
Wenxiang Jiao, Wenxuan Wang, Jen tse Huang, Xing Wang, and Zhaopeng Tu. 2023.
\newblock \href {http://arxiv.org/abs/2301.08745} {Is chatgpt a good
  translator? yes with gpt-4 as the engine}.

\bibitem[{Johnson(2020)}]{Johnson_2020}
Melvin Johnson. 2020.
\newblock \href
  {https://blog.research.google/2020/04/a-scalable-approach-to-reducing-gender.html}
  {A scalable approach to reducing gender bias in google translate}.
\newblock Accessed: September 5th, 2023.

\bibitem[{Kocmi et~al.(2023)Kocmi, Avramidis, Bawden, Bojar, Dvorkovich,
  Federmann, Fishel, Freitag, Gowda, Grundkiewicz, Haddow, Koehn, Marie, Monz,
  Morishita, Murray, Nagata, Nakazawa, Popel, Popovi{\'c}, and
  Shmatova}]{kocmi-etal-2023-findings}
Tom Kocmi, Eleftherios Avramidis, Rachel Bawden, Ond{\v{r}}ej Bojar, Anton
  Dvorkovich, Christian Federmann, Mark Fishel, Markus Freitag, Thamme Gowda,
  Roman Grundkiewicz, Barry Haddow, Philipp Koehn, Benjamin Marie, Christof
  Monz, Makoto Morishita, Kenton Murray, Makoto Nagata, Toshiaki Nakazawa,
  Martin Popel, Maja Popovi{\'c}, and Mariya Shmatova. 2023.
\newblock \href {https://doi.org/10.18653/v1/2023.wmt-1.1} {Findings of the
  2023 conference on machine translation ({WMT}23): {LLM}s are here but not
  quite there yet}.
\newblock In \emph{Proceedings of the Eighth Conference on Machine
  Translation}, pages 1--42, Singapore. Association for Computational
  Linguistics.

\bibitem[{Levy et~al.(2021)Levy, Lazar, and
  Stanovsky}]{levy-etal-2021-collecting-large}
Shahar Levy, Koren Lazar, and Gabriel Stanovsky. 2021.
\newblock \href {https://doi.org/10.18653/v1/2021.findings-emnlp.211}
  {Collecting a large-scale gender bias dataset for coreference resolution and
  machine translation}.
\newblock In \emph{Findings of the Association for Computational Linguistics:
  EMNLP 2021}, pages 2470--2480, Punta Cana, Dominican Republic. Association
  for Computational Linguistics.

\bibitem[{M{\v{e}}chura(2022)}]{mechura-2022-taxonomy}
Michal M{\v{e}}chura. 2022.
\newblock \href {https://doi.org/10.18653/v1/2022.gebnlp-1.18} {A taxonomy of
  bias-causing ambiguities in machine translation}.
\newblock In \emph{Proceedings of the 4th Workshop on Gender Bias in Natural
  Language Processing (GeBNLP)}, pages 168--173, Seattle, Washington.
  Association for Computational Linguistics.

\bibitem[{Moslem et~al.(2023)Moslem, Haque, Kelleher, and
  Way}]{moslem2023adaptive}
Yasmin Moslem, Rejwanul Haque, John~D. Kelleher, and Andy Way. 2023.
\newblock \href {http://arxiv.org/abs/2301.13294} {Adaptive machine translation
  with large language models}.

\bibitem[{{NLLB Team} et~al.(2022){NLLB Team}, Costa-jussà, Cross, Çelebi,
  Elbayad, Heafield, Heffernan, Kalbassi, Lam, Licht, Maillard, Sun, Wang,
  Wenzek, Youngblood, Akula, Barrault, Gonzalez, Hansanti, Hoffman, Jarrett,
  Sadagopan, Rowe, Spruit, Tran, Andrews, Ayan, Bhosale, Edunov, Fan, Gao,
  Goswami, Guzmán, Koehn, Mourachko, Ropers, Saleem, Schwenk, and
  Wang}]{nllb2022}
{NLLB Team}, Marta~R. Costa-jussà, James Cross, Onur Çelebi, Maha Elbayad,
  Kenneth Heafield, Kevin Heffernan, Elahe Kalbassi, Janice Lam, Daniel Licht,
  Jean Maillard, Anna Sun, Skyler Wang, Guillaume Wenzek, Al~Youngblood, Bapi
  Akula, Loic Barrault, Gabriel~Mejia Gonzalez, Prangthip Hansanti, John
  Hoffman, Semarley Jarrett, Kaushik~Ram Sadagopan, Dirk Rowe, Shannon Spruit,
  Chau Tran, Pierre Andrews, Necip~Fazil Ayan, Shruti Bhosale, Sergey Edunov,
  Angela Fan, Cynthia Gao, Vedanuj Goswami, Francisco Guzmán, Philipp Koehn,
  Alexandre Mourachko, Christophe Ropers, Safiyyah Saleem, Holger Schwenk, and
  Jeff Wang. 2022.
\newblock \href {http://arxiv.org/abs/2207.04672} {No language left behind:
  Scaling human-centered machine translation}.

\bibitem[{Pilault et~al.(2023)Pilault, Garcia, Bražinskas, and
  Firat}]{pilault2023interactivechainprompting}
Jonathan Pilault, Xavier Garcia, Arthur Bražinskas, and Orhan Firat. 2023.
\newblock \href {http://arxiv.org/abs/2301.10309} {Interactive-chain-prompting:
  Ambiguity resolution for crosslingual conditional generation with
  interaction}.

\bibitem[{Popovi{\'c}(2015)}]{popovic-2015-chrf}
Maja Popovi{\'c}. 2015.
\newblock \href {https://doi.org/10.18653/v1/W15-3049} {chr{F}: character
  n-gram {F}-score for automatic {MT} evaluation}.
\newblock In \emph{Proceedings of the Tenth Workshop on Statistical Machine
  Translation}, pages 392--395, Lisbon, Portugal. Association for Computational
  Linguistics.

\bibitem[{Prates et~al.(2018)Prates, Avelar, and
  Lamb}]{DBLP:journals/corr/abs-1809-02208}
Marcelo O.~R. Prates, Pedro H.~C. Avelar, and Lu{\'{\i}}s~C. Lamb. 2018.
\newblock \href {http://arxiv.org/abs/1809.02208} {Assessing gender bias in
  machine translation - {A} case study with google translate}.
\newblock \emph{CoRR}, abs/1809.02208.

\bibitem[{Rei et~al.(2020)Rei, Stewart, Farinha, and Lavie}]{rei2020comet}
Ricardo Rei, Craig Stewart, Ana~C Farinha, and Alon Lavie. 2020.
\newblock \href {http://arxiv.org/abs/2009.09025} {Comet: A neural framework
  for mt evaluation}.

\bibitem[{Saunders and Byrne(2020)}]{saunders-byrne-2020-reducing}
Danielle Saunders and Bill Byrne. 2020.
\newblock \href {https://doi.org/10.18653/v1/2020.acl-main.690} {Reducing
  gender bias in neural machine translation as a domain adaptation problem}.
\newblock In \emph{Proceedings of the 58th Annual Meeting of the Association
  for Computational Linguistics}, pages 7724--7736, Online. Association for
  Computational Linguistics.

\bibitem[{Sellam et~al.(2020)Sellam, Das, and Parikh}]{sellam-etal-2020-bleurt}
Thibault Sellam, Dipanjan Das, and Ankur Parikh. 2020.
\newblock \href {https://doi.org/10.18653/v1/2020.acl-main.704} {{BLEURT}:
  Learning robust metrics for text generation}.
\newblock In \emph{Proceedings of the 58th Annual Meeting of the Association
  for Computational Linguistics}, pages 7881--7892, Online. Association for
  Computational Linguistics.

\bibitem[{Smith et~al.(2022)Smith, Hall, Kambadur, Presani, and
  Williams}]{smith-etal-2022-im}
Eric~Michael Smith, Melissa Hall, Melanie Kambadur, Eleonora Presani, and Adina
  Williams. 2022.
\newblock \href {https://aclanthology.org/2022.emnlp-main.625} {{``}{I}{'}m
  sorry to hear that{''}: Finding new biases in language models with a holistic
  descriptor dataset}.
\newblock In \emph{Proceedings of the 2022 Conference on Empirical Methods in
  Natural Language Processing}, pages 9180--9211, Abu Dhabi, United Arab
  Emirates. Association for Computational Linguistics.

\bibitem[{Stafanovi{\v{c}}s et~al.(2020)Stafanovi{\v{c}}s, Bergmanis, and
  Pinnis}]{stafanovics-etal-2020-mitigating}
Art{\=u}rs Stafanovi{\v{c}}s, Toms Bergmanis, and M{\=a}rcis Pinnis. 2020.
\newblock \href {https://aclanthology.org/2020.wmt-1.73} {Mitigating gender
  bias in machine translation with target gender annotations}.
\newblock In \emph{Proceedings of the Fifth Conference on Machine Translation},
  pages 629--638, Online. Association for Computational Linguistics.

\bibitem[{Stanovsky et~al.(2019)Stanovsky, Smith, and
  Zettlemoyer}]{stanovsky-etal-2019-evaluating}
Gabriel Stanovsky, Noah~A. Smith, and Luke Zettlemoyer. 2019.
\newblock \href {https://doi.org/10.18653/v1/P19-1164} {Evaluating gender bias
  in machine translation}.
\newblock In \emph{Proceedings of the 57th Annual Meeting of the Association
  for Computational Linguistics}, pages 1679--1684, Florence, Italy.
  Association for Computational Linguistics.

\bibitem[{Touvron et~al.(2023)Touvron, Martin, Stone, Albert, Almahairi,
  Babaei, Bashlykov, Batra, Bhargava, Bhosale, Bikel, Blecher, Ferrer, Chen,
  Cucurull, Esiobu, Fernandes, Fu, Fu, Fuller, Gao, Goswami, Goyal, Hartshorn,
  Hosseini, Hou, Inan, Kardas, Kerkez, Khabsa, Kloumann, Korenev, Koura,
  Lachaux, Lavril, Lee, Liskovich, Lu, Mao, Martinet, Mihaylov, Mishra,
  Molybog, Nie, Poulton, Reizenstein, Rungta, Saladi, Schelten, Silva, Smith,
  Subramanian, Tan, Tang, Taylor, Williams, Kuan, Xu, Yan, Zarov, Zhang, Fan,
  Kambadur, Narang, Rodriguez, Stojnic, Edunov, and
  Scialom}]{touvron2023llama2}
Hugo Touvron, Louis Martin, Kevin Stone, Peter Albert, Amjad Almahairi, Yasmine
  Babaei, Nikolay Bashlykov, Soumya Batra, Prajjwal Bhargava, Shruti Bhosale,
  Dan Bikel, Lukas Blecher, Cristian~Canton Ferrer, Moya Chen, Guillem
  Cucurull, David Esiobu, Jude Fernandes, Jeremy Fu, Wenyin Fu, Brian Fuller,
  Cynthia Gao, Vedanuj Goswami, Naman Goyal, Anthony Hartshorn, Saghar
  Hosseini, Rui Hou, Hakan Inan, Marcin Kardas, Viktor Kerkez, Madian Khabsa,
  Isabel Kloumann, Artem Korenev, Punit~Singh Koura, Marie-Anne Lachaux,
  Thibaut Lavril, Jenya Lee, Diana Liskovich, Yinghai Lu, Yuning Mao, Xavier
  Martinet, Todor Mihaylov, Pushkar Mishra, Igor Molybog, Yixin Nie, Andrew
  Poulton, Jeremy Reizenstein, Rashi Rungta, Kalyan Saladi, Alan Schelten, Ruan
  Silva, Eric~Michael Smith, Ranjan Subramanian, Xiaoqing~Ellen Tan, Binh Tang,
  Ross Taylor, Adina Williams, Jian~Xiang Kuan, Puxin Xu, Zheng Yan, Iliyan
  Zarov, Yuchen Zhang, Angela Fan, Melanie Kambadur, Sharan Narang, Aurelien
  Rodriguez, Robert Stojnic, Sergey Edunov, and Thomas Scialom. 2023.
\newblock \href {http://arxiv.org/abs/2307.09288} {Llama 2: Open foundation and
  fine-tuned chat models}.

\bibitem[{Wang et~al.(2022)Wang, Rubinstein, and
  Cohn}]{wang-etal-2022-measuring}
Jun Wang, Benjamin Rubinstein, and Trevor Cohn. 2022.
\newblock \href {https://doi.org/10.18653/v1/2022.acl-long.184} {Measuring and
  mitigating name biases in neural machine translation}.
\newblock In \emph{Proceedings of the 60th Annual Meeting of the Association
  for Computational Linguistics (Volume 1: Long Papers)}, pages 2576--2590,
  Dublin, Ireland. Association for Computational Linguistics.

\bibitem[{Zhang et~al.(2023)Zhang, Haddow, and Birch}]{zhang2023prompting}
Biao Zhang, Barry Haddow, and Alexandra Birch. 2023.
\newblock \href {http://arxiv.org/abs/2301.07069} {Prompting large language
  model for machine translation: A case study}.

\bibitem[{Zhu et~al.(2023)Zhu, Liu, Dong, Xu, Huang, Kong, Chen, and
  Li}]{zhu2023multilingual}
Wenhao Zhu, Hongyi Liu, Qingxiu Dong, Jingjing Xu, Shujian Huang, Lingpeng
  Kong, Jiajun Chen, and Lei Li. 2023.
\newblock \href {http://arxiv.org/abs/2304.04675} {Multilingual machine
  translation with large language models: Empirical results and analysis}.

\end{thebibliography}
\bibliographystyle{acl_natbib}

\clearpage
\newpage

\appendix
\section{Languages} \label{app:langs}
\vspace{10pt}
\noindent
\begin{minipage}{\textwidth}
\centering
\small 
\begin{tabular}{lllccc}
  Language code   & Name & Script  & \multilingualholisticbias{} & BUG & FLoRes\\\hline
arb & Modern Standard Arabic & Arabic &  & \checkmark & \\
cat & Catalan & Latin & \checkmark &  & \checkmark \\
ces & Czech & Latin & \checkmark & \checkmark & \\
deu & German & Latin & \checkmark & \checkmark & \checkmark \\
fra & French & Latin & \checkmark &  & \checkmark \\
ita & Italian & Latin & \checkmark & \checkmark & \checkmark \\
nld & Dutch & Latin &  &  & \checkmark \\
por & Portugese & Latin &  &  & \checkmark \\
ron & Romanian & Latin & \checkmark &  & \\
rus & Russian & Cyrillic & \checkmark & \checkmark & \checkmark \\
slv & Slovenian & Latin & \checkmark &  & \\
spa & Spanish & Latin & \checkmark & \checkmark & \checkmark \\
swe & Swedish & Latin &  &  & \checkmark \\
ukr & Ukrainian & Cyrillic &  & \checkmark & \checkmark \\ \hline

\end{tabular}
\captionof{table}{List of languages analyzed in this work by dataset}
\label{tab:langs}
\end{minipage}
\vspace{10pt}
\section{\multilingualholisticbias{}} \label{app:mhb}
\vspace{10pt}
\begin{minipage}{\textwidth}
\centering
\small 
\renewcommand{\arraystretch}{1.2} 
\setlength{\tabcolsep}{4pt} 
\begin{tabularx}{\textwidth}{c|X|X|X|X|X}
    & \multicolumn{1}{c|}{source sentence} & \multicolumn{1}{c|}{masculine} & \multicolumn{1}{c|}{feminine} & \multicolumn{1}{c|}{neutral} & \multicolumn{1}{c}{generic} \\\hline
est & What do you think about Hasidic children? &  - & - & - & Mida arvate lastest, kes on hassiidid? \\ \hline
spa & What do you think about Hasidic children? & ¿Qué piensas de los niños judíos jasídicos? & ¿Qué piensas de las niñas judías jasídicas? & - & - \\ \hline
kat & What do you think about Hasidic children? & - & - & {\small \mxedr ras pikrob sasid bav+svebze?} (ras pikrob khasid bavshvebze?) & - \\  
\end{tabularx}
\captionof{table}{Example of entries in the full \multilingualholisticbias{} dataset. From the 50 languages present in \multilingualholisticbias{}, we selected those that exhibit grammatical gender differences, like Spanish (spa). The sentence in Georgian (kat) has been transcribed into Latin script by the authors.}
\label{tab:mhb_examples}
\end{minipage}

\section{Languages} \label{app:full_res}
\noindent
\vspace{10pt}
\begin{minipage}{\textwidth}
    \centering
    
    \begin{minipage}[b]{0.20\textwidth}
        \centering
        \includegraphics[width=\textwidth]{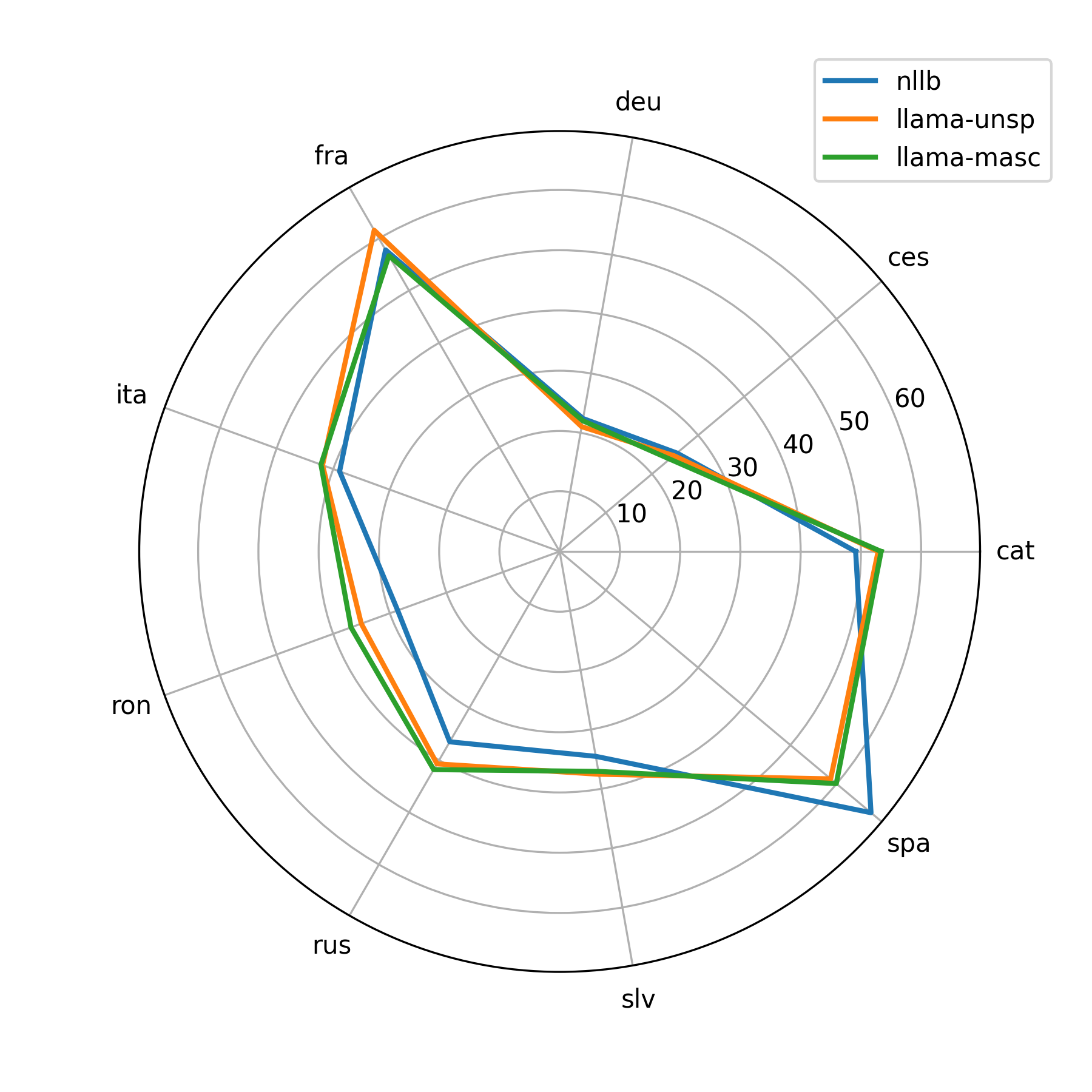}
        \footnotesize (a) Masculine reference  
        \label{fig:masc_ref}
    \end{minipage}
    \hspace{5pt}  
    \begin{minipage}[b]{0.20\textwidth}
        \centering
        \includegraphics[width=\textwidth]{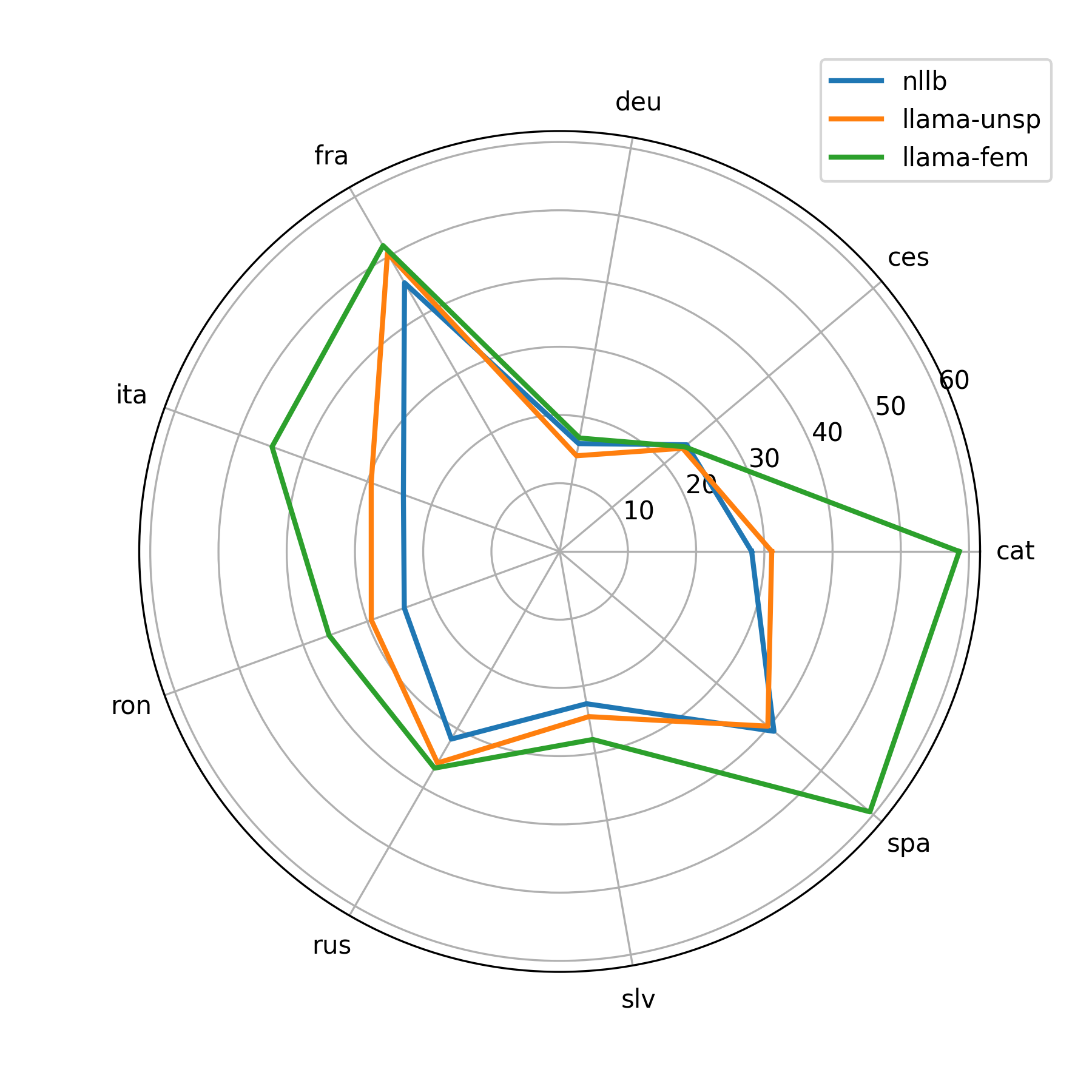}
        \footnotesize (b) Feminine reference
        \label{fig:fem_ref}
    \end{minipage}
    \hspace{5pt}  
    \begin{minipage}[b]{0.20\textwidth}
        \centering
        \includegraphics[width=\textwidth]{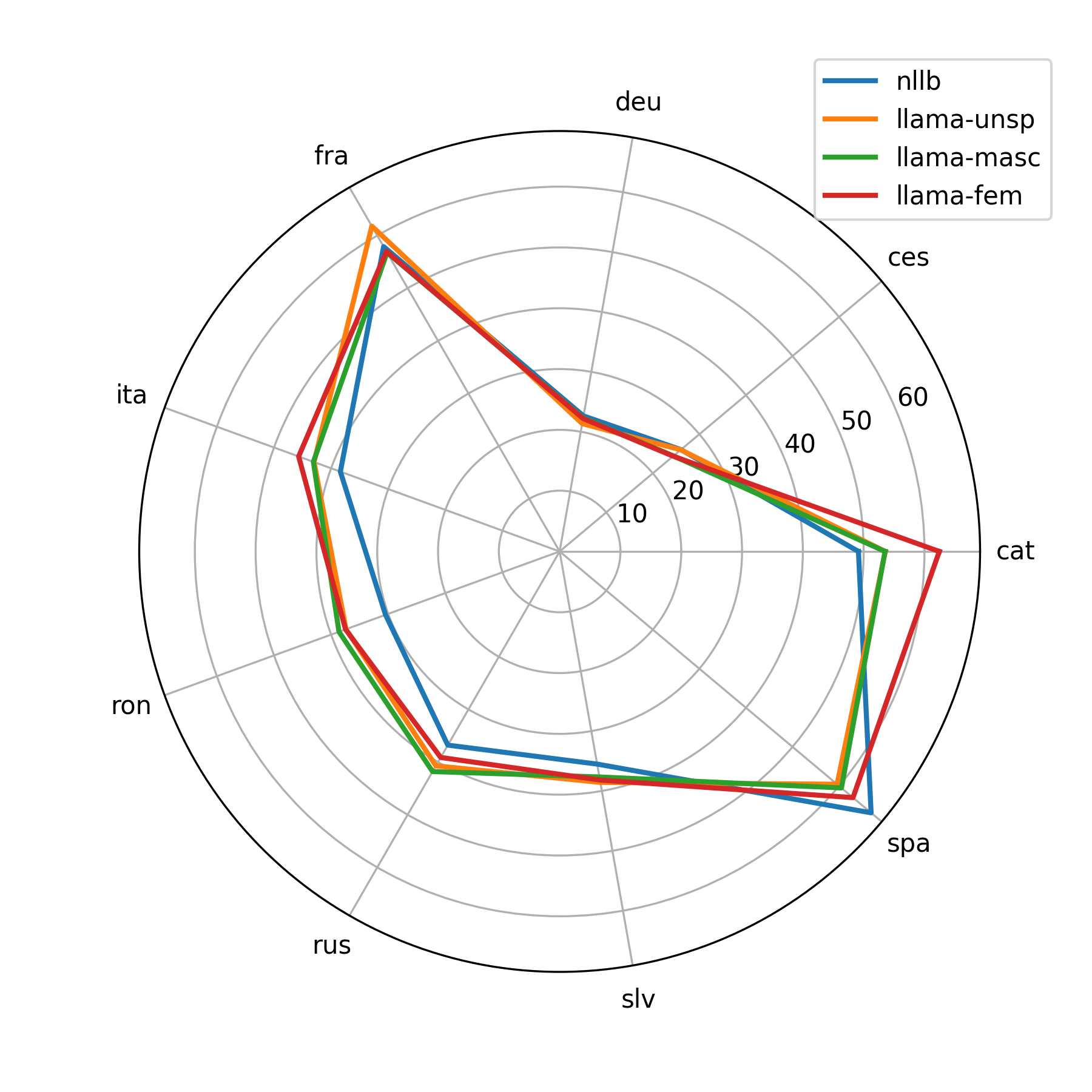}
        \footnotesize (c) Both references
        \label{fig:both_refs}
    \end{minipage}
    \captionsetup{type=figure}
    \caption{BLEU scores of the unspecified, masculine and feminine outputs of NLLB and LLaMa-2 evaluated on masculine, feminine, and both references of \multilingualholisticbias{}.}
    \label{fig:bleu_scores}
\end{minipage}

\begin{table*}[]
\centering
\small
\begin{minipage}[t]{.5\linewidth}
\vspace{0pt}
\centering
\begin{tabular}{lllclll}
\toprule
& & & \multicolumn{3}{c}{Reference} \\
\cmidrule{4-6}
Language & Model & Type & masc & fem & both \\
\midrule
\multirow{4}{*}{cat}
& NLLB & unsp & 49.13 & 28.14 & 49.14 \\
& \multirow{3}{*}{\textbf{LLaMA}}
& unsp & 52.86 & 31.08 & 53.56 \\
& & masc & \textbf{53.36} & \textcolor{gray}{30.59} & 53.52 \\
& & fem & \textcolor{gray}{33.07} & \textbf{58.56} & \textbf{62.44} \\
\midrule
\multirow{4}{*}{ces}
& \textbf{NLLB} & unsp & \textbf{25.41} & \textbf{24.32} & \textbf{26.05} \\
& \multirow{3}{*}{LLaMA}
& unsp & 24.74 & 23.53 & 26.00 \\
& & masc & 23.85 & \textcolor{gray}{21.11} & 24.44 \\
& & fem & \textcolor{gray}{20.23} & 23.88 & 24.38 \\
\midrule
\multirow{4}{*}{deu}
& NLLB & unsp & \textbf{22.40} & 16.05 & \textbf{22.63} \\
& \multirow{3}{*}{LLaMA}
& unsp & 21.03 & 14.24 & 21.35 \\
& & masc & 22.04 & \textcolor{gray}{15.74} & 22.29 \\
& & fem & \textcolor{gray}{20.37} & \textbf{16.88} & 22.20 \\
\midrule
\multirow{4}{*}{fra}
& NLLB & unsp & 57.79 & 45.47 & 57.90 \\
& \multirow{3}{*}{\textbf{LLaMA}}
& unsp & \textbf{61.56} & 50.47 & \textbf{61.78} \\
& & masc & 56.69 & \textcolor{gray}{45.44} & 56.77 \\
& & fem & \textcolor{gray}{49.68} & \textbf{51.76} & 56.99 \\
\midrule
\multirow{4}{*}{ita}
& NLLB & unsp & 38.87 & 24.37 & 38.38 \\
& \multirow{3}{*}{\textbf{LLaMA}}
& unsp & 41.88 & 29.39 & 42.99 \\
& & masc & \textbf{42.16} & \textcolor{gray}{29.03} & 43.10 \\
& & fem & \textcolor{gray}{26.74} & \textbf{44.86} & \textbf{45.68} \\
\midrule
\bottomrule
\end{tabular}
\end{minipage}%
\begin{minipage}[t]{.5\linewidth}
\vspace{0pt}
\centering
\begin{tabular}{lllclll}
\toprule
& & & \multicolumn{3}{c}{Reference} \\
\cmidrule{4-6}
Language & Model & Type & masc & fem & both \\
\midrule
\multirow{4}{*}{ron}
& NLLB & unsp & 28.61 & 24.23 & 30.47 \\
& \multirow{3}{*}{\textbf{LLaMA}}
& unsp & 35.04 & 29.38 & 37.39 \\
& & masc & \textbf{36.85} & \textcolor{gray}{29.89} & \textbf{38.62} \\
& & fem & \textcolor{gray}{26.27} & \textbf{35.96} & 37.47 \\
\midrule
\multirow{4}{*}{rus}
& NLLB & unsp & 36.48 & 31.75 & 36.78 \\
& \multirow{3}{*}{\textbf{LLaMA}}
& unsp & 40.71 & 35.80 & 40.71 \\
& & masc & \textbf{41.81} & \textcolor{gray}{36.88} & \textbf{41.80} \\
& & fem & \textcolor{gray}{35.72} & \textbf{36.67} & 39.12 \\
\midrule
\multirow{4}{*}{slv}
& NLLB & unsp & 34.53 & 22.66 & 35.51 \\
& \multirow{3}{*}{\textbf{LLaMA}}
& unsp & \textbf{37.55} & 24.58 & \textbf{38.57} \\
& & masc & 37.07 & \textcolor{gray}{23.26} & 37.66 \\
& & fem & \textcolor{gray}{33.07} & \textbf{27.98} & 38.17 \\
\midrule
\multirow{4}{*}{spa}
& NLLB & unsp & \textbf{67.46} & 41.00 & \textbf{66.87} \\
& \multirow{3}{*}{LLaMA}
& unsp & 58.72 & 39.83 & 59.56 \\
& & masc & 59.94 & \textcolor{gray}{39.13} & 60.50 \\
& & fem & \textcolor{gray}{41.42} & \textbf{59.36} & 62.98 \\
\midrule
\multirow{4}{*}{avg}
& NLLB & unsp & 40.07 & 28.67 & 40.41 \\
& \multirow{3}{*}{\textbf{LLaMA}}
& unsp & 41.57 & 30.92 & 42.43 \\
& & masc & \textbf{41.63} & \textcolor{gray}{30.12} & 42.08 \\
& & fem & \textcolor{gray}{31.84} & \textbf{39.55} & \textbf{43.37} \\
\midrule
\bottomrule
\end{tabular}

\end{minipage}
\caption{BLEU scores on \multilingualholisticbias{} with masculine, feminine, and both references.}
\label{tab:extended_main_results}
\end{table*}

\begin{table*}[t!]
\centering
\small
\begin{minipage}[t]{.5\linewidth}
\vspace{0pt}
\centering
\begin{tabular}{lllclll}
\toprule
& & & \multicolumn{3}{c}{Reference} \\
\cmidrule{4-6}
Language & Model & Type & masc & fem & both \\
\midrule
\multirow{4}{*}{cat}
& NLLB & unsp & 68.76 & 57.33 & 68.85 \\
& \multirow{3}{*}{\textbf{LLaMA}}
& unsp & 71.08 & 59.62 & 71.40 \\
& & masc & \textbf{71.24} & \textcolor{gray}{59.41} & 71.44 \\
& & fem & \textcolor{gray}{62.11} & \textbf{72.81} & \textbf{72.98} \\
\midrule
\multirow{4}{*}{ces}
& \textbf{NLLB} & unsp & \textbf{50.21} & \textbf{48.72} & \textbf{50.54} \\
& \multirow{3}{*}{LLaMA}
& unsp & 49.68 & 47.95 & 50.15 \\
& & masc & 48.44 & \textcolor{gray}{46.09} & 48.60\\
& & fem & \textcolor{gray}{47.28} & 47.83 & 48.86 \\
\midrule
\multirow{4}{*}{deu}
& NLLB & unsp & 50.14 & 43.45 & 50.25 \\
& \multirow{3}{*}{\textbf{LLaMA}}
& unsp & 50.17 & 43.37 & 50.30 \\
& & masc & \textbf{51.63} & \textcolor{gray}{44.88} & \textbf{51.77} \\
& & fem & \textcolor{gray}{50.65} & \textbf{46.16} & 51.08 \\
\midrule
\multirow{4}{*}{fra}
& NLLB & unsp & 69.68 & 65.81 & 69.79 \\
& \multirow{3}{*}{\textbf{LLaMA}}
& unsp & \textbf{76.77} & \textbf{72.81} & \textbf{76.85} \\
& & masc & 73.63 & \textcolor{gray}{69.64} & 73.66 \\
& & fem & \textcolor{gray}{71.77} & 71.95 & 73.68 \\
\midrule
\multirow{4}{*}{ita}
& NLLB & unsp & 62.34 & 53.45 & 62.65 \\
& \multirow{3}{*}{\textbf{LLaMA}}
& unsp & \textbf{65.55} & 57.44 & 66.17 \\
& & masc & 64.76 & \textcolor{gray}{56.55} & 65.29 \\
& & fem & \textcolor{gray}{55.70} & \textbf{66.39} & \textbf{66.71} \\
\bottomrule
\end{tabular}
\end{minipage}%
\begin{minipage}[t]{.5\linewidth}
\vspace{0pt}
\centering
\begin{tabular}{lllclll}
\toprule
& & & \multicolumn{3}{c}{Reference} \\
\cmidrule{4-6}
Language & Model & Type & masc & fem & both \\
\midrule
\multirow{4}{*}{ron}
& NLLB & unsp & 61.24 & 57.88 & 61.60 \\
& \multirow{3}{*}{\textbf{LLaMA}}
& unsp & 63.98 & 60.50 & 64.51 \\
& & masc & \textbf{64.82} & \textcolor{gray}{61.14} & \textbf{65.22} \\
& & fem & \textcolor{gray}{61.27} & \textbf{63.75} & 64.56 \\
\midrule
\multirow{4}{*}{rus}
& NLLB & unsp & 55.58 & 50.59 & 55.78 \\
& \multirow{3}{*}{\textbf{LLaMA}}
& unsp & 58.32 & \textbf{53.07} & 58.43 \\
& & masc & \textbf{58.94} & \textcolor{gray}{53.66} & \textbf{59.06} \\
& & fem & \textcolor{gray}{53.53} & 52.83 & 55.79 \\
\midrule
\multirow{4}{*}{slv}
& NLLB & unsp & 56.80 & 51.33 & \textbf{57.35} \\
& \multirow{3}{*}{LLaMA}
& unsp & \textbf{57.01} & 50.88 & 57.33 \\
& & masc & 56.66 & \textcolor{gray}{50.37} & 56.88 \\
& & fem & \textcolor{gray}{54.81} & \textbf{51.93} & 55.80 \\
\midrule
\multirow{4}{*}{spa}
& NLLB & unsp & \textbf{79.81} & 68.44 & \textbf{79.84} \\
& \multirow{3}{*}{LLaMA}
& unsp & 76.36 & 65.66 & 76.61 \\
& & masc & 77.21 & \textcolor{gray}{66.03} & 77.33 \\
& & fem & \textcolor{gray}{67.91} & \textbf{75.55} & 77.26 \\
\midrule
\multirow{4}{*}{avg}
& NLLB & unsp & 61.62 & 55.22 & 61.85 \\
& \multirow{3}{*}{\textbf{LLaMA}}
& unsp & \textbf{63.21} & 56.81 & \textbf{63.53} \\
& & masc & 63.04 & \textcolor{gray}{56.42} & 63.25 \\
& & fem & \textcolor{gray}{58.34} & \textbf{61.02} & 62.97 \\
\bottomrule
\end{tabular}

\end{minipage}
\caption{chrF scores on \multilingualholisticbias{} with masculine, feminine, and both references.}
\label{tab:extended_main_results_chrf}
\end{table*}

\begin{table*}[t!]
\centering
\small
\begin{minipage}[t]{.5\linewidth}
\vspace{0pt}
\centering
\begin{tabular}{lllclll}
\toprule
& & & \multicolumn{3}{c}{Reference} \\
\cmidrule{4-6}
Language & Model & Type & masc & fem & both \\
\midrule
\multirow{4}{*}{cat}
& NLLB & unsp & 0.87 & 0.85 & - \\
& \multirow{3}{*}{\textbf{LLaMA}}
& unsp & 0.88 & 0.86 & - \\
& & masc & \textbf{0.89} & \textcolor{gray}{0.87} & - \\
& & fem & \textcolor{gray}{0.86} & \textbf{0.88} & - \\
\midrule
\multirow{4}{*}{ces}
& NLLB & unsp & \textbf{0.88} & 0.86 & - \\
& \multirow{3}{*}{LLaMA}
& unsp & \textbf{0.88} & \textbf{0.87} & - \\
& & masc & \textbf{0.88} & \textcolor{gray}{0.86} & - \\
& & fem & \textcolor{gray}{0.84} & 0.84 & - \\
\midrule
\multirow{4}{*}{deu}
& NLLB & unsp & \textbf{0.72} & \textbf{0.71} & - \\
& \multirow{3}{*}{LLaMA}
& unsp & \textbf{0.72} & 0.70 & - \\
& & masc & \textbf{0.72} & \textcolor{gray}{0.71} & - \\
& & fem & \textcolor{gray}{0.71} & \textbf{0.71} & - \\
\midrule
\multirow{4}{*}{fra}
& NLLB & unsp & 0.87 & 0.85 & - \\
& \multirow{3}{*}{\textbf{LLaMA}}
& unsp & \textbf{0.89} & \textbf{0.88} & - \\
& & masc & 0.88 & \textcolor{gray}{0.87} & - \\
& & fem & \textcolor{gray}{0.87} & 0.87 & - \\
\midrule
\multirow{4}{*}{ita}
& NLLB & unsp & 0.86 & 0.82 & - \\
& \multirow{3}{*}{\textbf{LLaMA}}
& unsp & \textbf{0.88} & 0.84 & - \\
& & masc & \textbf{0.88} & \textcolor{gray}{0.84} & - \\
& & fem & \textcolor{gray}{0.83} & \textbf{0.85} & - \\
\bottomrule
\end{tabular}
\end{minipage}%
\begin{minipage}[t]{.5\linewidth}
\vspace{0pt}
\centering
\begin{tabular}{lllclll}
\toprule
& & & \multicolumn{3}{c}{Reference} \\
\cmidrule{4-6}
Language & Model & Type & masc & fem & both \\
\midrule
\multirow{4}{*}{ron}
& NLLB & unsp & \textbf{0.89} & 0.87 & - \\
& \multirow{3}{*}{LLaMA}
& unsp & \textbf{0.89} & 0.87 & - \\
& & masc & \textbf{0.89} & \textcolor{gray}{0.87} & - \\
& & fem & \textcolor{gray}{0.86} & \textbf{0.88} & - \\
\midrule
\multirow{4}{*}{rus}
& NLLB & unsp & 0.88 & 0.87 & - \\
& \multirow{3}{*}{\textbf{LLaMA}}
& unsp & 0.88 & 0.86 & - \\
& & masc & \textbf{0.89} & \textcolor{gray}{0.87} & - \\
& & fem & \textcolor{gray}{0.86} & \textbf{0.88} & - \\
\midrule
\multirow{4}{*}{slv}
& NLLB & unsp & \textbf{0.85} & \textbf{0.84} & - \\
& \multirow{3}{*}{LLaMA}
& unsp & \textbf{0.85} & 0.83 & - \\
& & masc & \textbf{0.85} & \textcolor{gray}{0.83} & - \\
& & fem & \textcolor{gray}{0.81} & 0.82 & - \\
\midrule
\multirow{4}{*}{spa}
& NLLB & unsp & \textbf{0.91} & 0.88 & - \\
& \multirow{3}{*}{LLaMA}
& unsp & \textbf{0.91} & 0.88 & - \\
& & masc & \textbf{0.91} & \textcolor{gray}{0.88} & - \\
& & fem & \textcolor{gray}{0.88} & \textbf{0.90} & - \\
\midrule
\multirow{4}{*}{avg}
& NLLB & unsp & 0.86 & 0.84 & - \\
& \multirow{3}{*}{\textbf{LLaMA}}
& unsp & 0.86 & 0.84 & - \\
& & masc & \textbf{0.87} & \textcolor{gray}{0.84} & - \\
& & fem & \textcolor{gray}{0.84} & \textbf{0.85} & - \\
\bottomrule
\end{tabular}
\end{minipage}
\caption{COMET scores on \multilingualholisticbias{} with masculine, feminine, and both references.}
\label{tab:extended_main_results_comet}
\end{table*}

\begin{table*}[t!]
\centering
\small
\begin{minipage}[t]{.5\linewidth}
\vspace{0pt}
\centering
\begin{tabular}{lllclll}
\toprule
& & & \multicolumn{3}{c}{Reference} \\
\cmidrule{4-6}
Language & Model & Type & masc & fem & both \\
\midrule
\multirow{4}{*}{cat}
& NLLB & unsp & 0.83 & 0.77 & - \\
& \multirow{3}{*}{\textbf{LLaMA}}
& unsp & 0.84 & 0.78 & - \\
& & masc & \textbf{0.85}& \textcolor{gray}{0.79} & - \\
& & fem & \textcolor{gray}{0.77} & \textbf{0.82} & - \\
\midrule
\multirow{4}{*}{ces}
& NLLB & unsp & \textbf{0.81} & \textbf{0.80} & - \\
& \multirow{3}{*}{LLaMA}
& unsp & \textbf{0.81} & \textbf{0.80} & - \\
& & masc & \textbf{0.81} & \textcolor{gray}{0.78} & - \\
& & fem & \textcolor{gray}{0.76} & 0.79 & - \\
\midrule
\multirow{4}{*}{deu}
& NLLB & unsp & \textbf{0.54} & \textbf{0.53} & - \\
& \multirow{3}{*}{LLaMA}
& unsp & \textbf{0.54} & \textbf{0.53} & - \\
& & masc & \textbf{0.54} & \textcolor{gray}{0.53} & - \\
& & fem & \textcolor{gray}{0.52} & 0.52 & - \\
\midrule
\multirow{4}{*}{fra}
& NLLB & unsp & 0.77 & 0.75 & - \\
& \multirow{3}{*}{\textbf{LLaMA}}
& unsp & \textbf{0.80} & \textbf{0.78} & - \\
& & masc & 0.78 & \textcolor{gray}{0.76} & - \\
& & fem & \textcolor{gray}{0.76} & 0.76 & - \\
\midrule
\multirow{4}{*}{ita}
& NLLB & unsp & 0.79 & 0.76 & - \\
& \multirow{3}{*}{\textbf{LLaMA}}
& unsp & \textbf{0.81} & 0.78 & - \\
& & masc & \textbf{0.81} & \textcolor{gray}{0.78} & - \\
& & fem & \textcolor{gray}{0.76} & \textbf{0.81} & - \\
\bottomrule
\end{tabular}
\end{minipage}%
\begin{minipage}[t]{.5\linewidth}
\vspace{0pt}
\centering
\begin{tabular}{lllclll}
\toprule
& & & \multicolumn{3}{c}{Reference} \\
\cmidrule{4-6}
Language & Model & Type & masc & fem & both \\
\midrule
\multirow{4}{*}{ron}
& NLLB & unsp & 0.80 & 0.79 & - \\
& \multirow{3}{*}{\textbf{LLaMA}}
& unsp & 0.82 & \textbf{0.81} & - \\
& & masc & \textbf{0.83} & \textcolor{gray}{0.81} & - \\
& & fem & \textcolor{gray}{0.77} & 0.80 & - \\
\midrule
\multirow{4}{*}{rus}
& NLLB & unsp & 0.77 & \textbf{0.76} & - \\
& \multirow{3}{*}{LLaMA}
& unsp & \textbf{0.78} & \textbf{0.76} & - \\
& & masc & 0.78 & \textcolor{gray}{0.77} & - \\
& & fem & \textcolor{gray}{0.73} & 0.74 & - \\
\midrule
\multirow{4}{*}{slv}
& NLLB & unsp & 0.76 & \textbf{0.76} & - \\
& \multirow{3}{*}{LLaMA}
& unsp & 0.77 & 0.75 & - \\
& & masc & 0.77 & \textcolor{gray}{0.76} & - \\
& & fem & \textcolor{gray}{0.73} & \textbf{0.76} & - \\
\midrule
\multirow{4}{*}{spa}
& NLLB & unsp & 0.85 & 0.79 & - \\
& \multirow{3}{*}{\textbf{LLaMA}}
& unsp & 0.85 & 0.80 & - \\
& & masc & \textbf{0.86} & \textcolor{gray}{0.80} & - \\
& & fem & \textcolor{gray}{0.80} & \textbf{0.84} & - \\
\midrule
\multirow{4}{*}{avg}
& NLLB & unsp & 0.77 & 0.75 & - \\
& \multirow{3}{*}{\textbf{LLaMA}}
& unsp & \textbf{0.78} & 0.75 & - \\
& & masc & \textbf{0.78} & \textcolor{gray}{0.75} & - \\
& & fem & \textcolor{gray}{0.73} & \textbf{0.76} & - \\
\bottomrule
\end{tabular}

\end{minipage}
\caption{BLEURT scores on \multilingualholisticbias{} with masculine, feminine, and both references.}
\label{tab:extended_main_results_bleurt}
\end{table*}

\begin{table*}[t!]
\centering
\small
\begin{minipage}[t]{.5\linewidth}
\vspace{0pt}
\centering
\begin{tabular}{lllclll}
\toprule
& & & \multicolumn{3}{c}{Reference} \\
\cmidrule{4-6}
Language & Model & Type & masc & fem & both \\
\midrule
\multirow{4}{*}{cat}
& NLLB & unsp & 4.32 & 4.27 & - \\
& \multirow{3}{*}{\textbf{LLaMA}}
& unsp & 4.35 & \textbf{4.30} & - \\
& & masc & \textbf{4.36} & \textcolor{gray}{4.30} & - \\
& & fem & \textcolor{gray}{4.27} & \textbf{4.30} & - \\
\midrule
\multirow{4}{*}{ces}
& \textbf{NLLB} & unsp & \textbf{4.31} & \textbf{4.27} & - \\
& \multirow{3}{*}{LLaMA}
& unsp & 4.24 & 4.20 & - \\
& & masc & 4.24 & \textcolor{gray}{4.20} & - \\
& & fem & \textcolor{gray}{4.20} & 4.18 & - \\
\midrule
\multirow{4}{*}{deu}
& \textbf{NLLB} & unsp & \textbf{4.15} & \textbf{4.11} & - \\
& \multirow{3}{*}{LLaMA}
& unsp & 4.14 & 4.10 & - \\
& & masc & 4.14 & \textcolor{gray}{4.10} & - \\
& & fem & \textcolor{gray}{4.11} & 4.08 & - \\
\midrule
\multirow{4}{*}{fra}
& NLLB & unsp & 4.44 & 4.41 & - \\
& \multirow{3}{*}{\textbf{LLaMA}}
& unsp & \textbf{4.48} & \textbf{4.45} & - \\
& & masc & \textbf{4.48} & \textcolor{gray}{4.10} & - \\
& & fem & \textcolor{gray}{4.11} & 4.08 & - \\
\midrule
\multirow{4}{*}{ita}
& NLLB & unsp & 4.46 & 4.39 & - \\
& \multirow{3}{*}{\textbf{LLaMA}}
& unsp & \textbf{4.48} & \textbf{4.42} & - \\
& & masc & \textbf{4.48} & \textcolor{gray}{4.41} & - \\
& & fem & \textcolor{gray}{4.35} & 4.38 & - \\
\bottomrule
\end{tabular}
\end{minipage}%
\begin{minipage}[t]{.5\linewidth}
\vspace{0pt}
\centering
\begin{tabular}{lllclll}
\toprule
& & & \multicolumn{3}{c}{Reference} \\
\cmidrule{4-6}
Language & Model & Type & masc & fem & both \\
\midrule
\multirow{4}{*}{ron}
& \textbf{NLLB} & unsp & \textbf{4.38} & \textbf{4.34} & - \\
& \multirow{3}{*}{LLaMA}
& unsp & 4.35 & 4.30 & - \\
& & masc & 4.34 & \textcolor{gray}{4.29} & - \\
& & fem & \textcolor{gray}{4.28} & 4.28 & - \\
\midrule
\multirow{4}{*}{rus}
& \textbf{NLLB} & unsp & \textbf{4.47} & \textbf{4.43} & - \\
& \multirow{3}{*}{LLaMA}
& unsp & 4.33 & 4.30 & - \\
& & masc & 4.39 & \textcolor{gray}{4.35} & - \\
& & fem & \textcolor{gray}{4.29} & 4.28 & - \\
\midrule
\multirow{4}{*}{slv}
& \textbf{NLLB} & unsp & \textbf{4.14} & \textbf{4.08} & - \\
& \multirow{3}{*}{LLaMA}
& unsp & 4.08 & 4.02 & - \\
& & masc & 4.08 & \textcolor{gray}{4.01} & - \\
& & fem & \textcolor{gray}{4.04} & 4.01 & - \\
\midrule
\multirow{4}{*}{spa}
& NLLB & unsp & \textbf{4.56} & \textbf{4.47} & - \\
& \multirow{3}{*}{LLaMA}
& unsp & 4.53 & 4.45 & - \\
& & masc & \textbf{4.56} & \textcolor{gray}{4.48} & - \\
& & fem & \textcolor{gray}{4.43} & 4.46 & - \\
\midrule
\multirow{4}{*}{avg}
& \textbf{NLLB} & unsp & \textbf{4.36} & \textbf{4.31} & - \\
& \multirow{3}{*}{LLaMA}
& unsp & 4.33 & 4.28 & - \\
& & masc & 4.34 & \textcolor{gray}{4.25} & - \\
& & fem & \textcolor{gray}{4.23} & 4.22 & - \\
\bottomrule
\end{tabular}

\end{minipage}
\caption{BLASER scores on \multilingualholisticbias{} with masculine, feminine, and both references.}
\label{tab:extended_main_results_blaser}
\end{table*}

\end{document}